\documentclass[twoside,leqno,twocolumn]{article}
\usepackage[letterpaper]{geometry}
\usepackage{siamproceedings}

\usepackage[T1]{fontenc}
\usepackage{amsfonts}
\usepackage{graphicx}
\usepackage{epstopdf}
\usepackage{enumitem}
\usepackage{algorithmic}
\ifpdf
  \DeclareGraphicsExtensions{.eps,.pdf,.png,.jpg}
\else
  \DeclareGraphicsExtensions{.eps}
\fi

\newsiamremark{remark}{Remark}
\newsiamremark{hypothesis}{Hypothesis}
\crefname{hypothesis}{Hypothesis}{Hypotheses}
\newsiamthm{claim}{Claim}

\usepackage{amsopn}

\usepackage{colortbl}
\usepackage{multirow}
\usepackage{booktabs}
\usepackage{subcaption}
\usepackage{float}
\usepackage{xcolor} 
\usepackage{placeins}
\usepackage{dblfloatfix}
\begin{document}

\title{\Large ACT: Anti-Crosstalk Learning for Cross-Sectional Stock Ranking via Temporal Disentanglement and Structural Purification}
    \author{Juntao Li\thanks{ School of Computing and Data Science, University of Hong Kong, Hong Kong SAR. Email: juntao.li@connect.hku.hk}
    \and Liang Zhang \thanks{Thrust of Financial Technology, Hong Kong University of Science and Technology (Guangzhou), China. Email: liangzhang@hkust-gz.edu.cn} \thanks{Corresponding author.}}
\date{}

\maketitle
\fancyfoot[R]{\scriptsize{Copyright \textcopyright\ 2026 by SIAM\\
Unauthorized reproduction of this article is prohibited}}

\begin{abstract}
Cross-sectional stock ranking is a fundamental task in quantitative investment, relying on both temporal modeling of individual stocks and the capture of inter-stock dependencies. While existing deep learning models leverage graph-based approaches to enhance ranking accuracy by propagating information over relational graphs, they suffer from a key challenge: crosstalk, namely unintended information interference across predictive factors. We identify two forms of crosstalk: temporal-scale crosstalk, where trends, fluctuations, and shocks are entangled in a shared representation and non-transferable local patterns contaminate cross-stock learning; and structural crosstalk, where heterogeneous relations are indiscriminately fused and relation-specific predictive signals are obscured. To address both issues, we propose the Anti-CrossTalk (ACT) framework for cross-sectional stock ranking via temporal disentanglement and structural purification. Specifically, ACT first decomposes each stock sequence into trend, fluctuation, and shock components, then extracts component-specific information through dedicated branches, which effectively decouples non-transferable local patterns. ACT further introduces a Progressive Structural Purification Encoder to sequentially purify structural crosstalk on the trend component after mitigating temporal-scale crosstalk. An adaptive fusion module finally integrates all branch representations for ranking. Experiments on CSI\,300 and CSI\,500 demonstrate that ACT achieves state-of-the-art ranking accuracy and superior portfolio performance, with improvements of up to \textbf{74.25\%} on the CSI300 dataset. 

\end{abstract}

\section{Introduction.}

Cross-sectional stock ranking aims to predict the relative return ordering of stocks at each trading day, serving as a fundamental task in quantitative investment~\cite{Gu23,Li24,Sonkavde23,Patel25}. Early approaches rely on hand-crafted indicators such as moving averages~\cite{Babu14} and price-to-earnings ratios~\cite{Lewellen02}, which are often limited in capturing complex market dynamics. To address this, deep learning models have been introduced to learn representations directly from data. Recurrent architectures such as LSTM~\cite{Hochreiter97} and GRU~\cite{cho2014gru}, as well as Transformer-based models~\cite{vaswani2023attention,enhancetrans19}, have demonstrated strong capability in modeling temporal dependencies, while time-series decomposition~\cite{Chengyunyao23} further improves predictive performance by decomposing underlying patterns. Beyond temporal modeling, recent work emphasizes the importance of explicitly capturing inter-stock relationships. Graph neural networks (GNNs), including GCN~\cite{Kipf17} and GAT~\cite{Velickovic17,Niu25}, construct relational graphs to encode dependencies among stocks, while spatiotemporal graph models integrate both temporal and spatial interactions~\cite{Shao22,Ma23,Wang20}. 

\begin{figure}[t]
    \centering
    \includegraphics[width=0.40\textwidth]{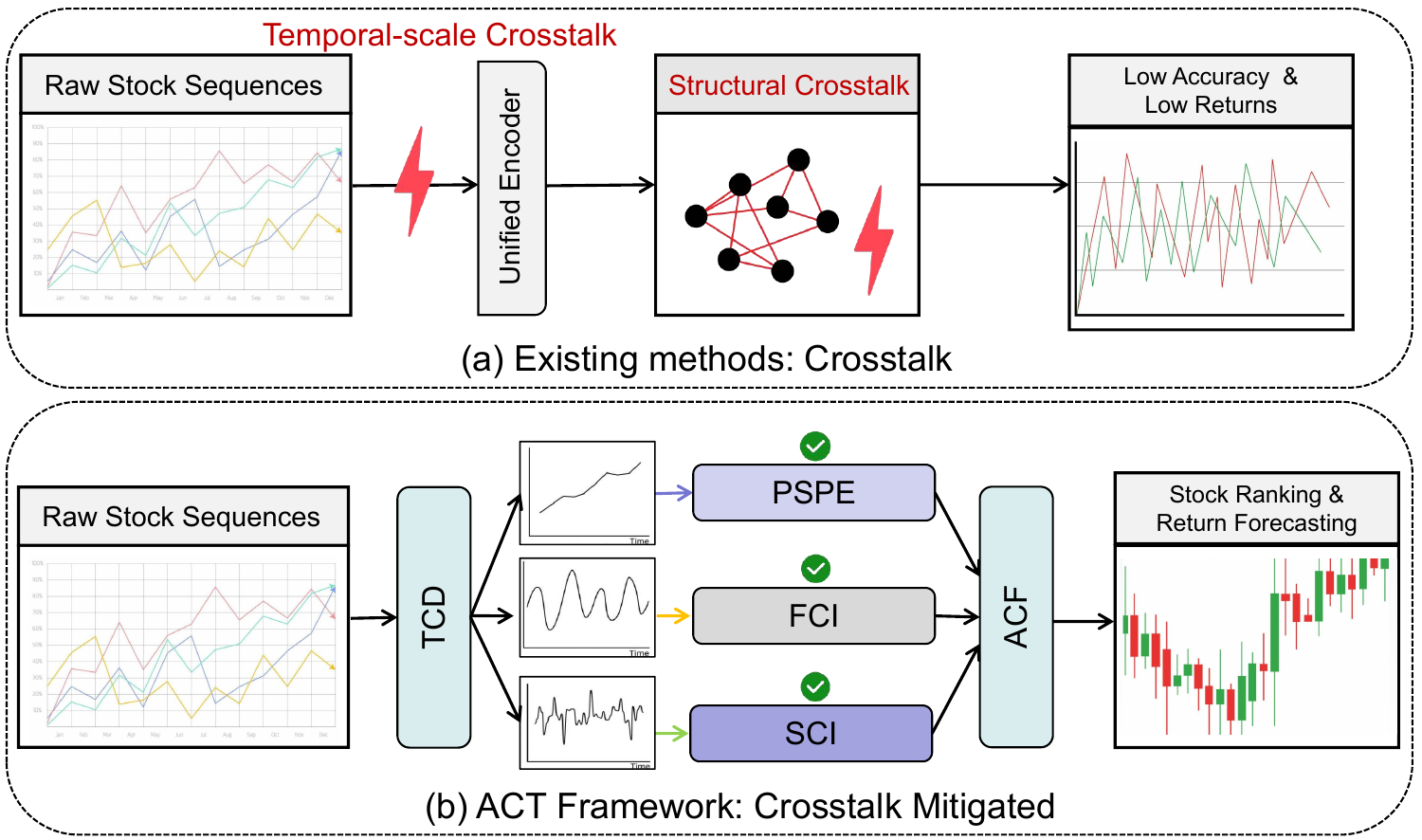}
    \caption{ (a) Existing methods induce crosstalk by entangling heterogeneous temporal dynamics and relational structures within unified representations. (b) The ACT framework mitigates leakage via TCD-based disentanglement and dedicated pathway encoding (PSPE/FCI/SCI) before adaptive fusion (ACF).}
    \label{fig:framework1}
    \vspace{-0.4em}
\end{figure}

Despite these advances, existing graph-based models still suffer from a fundamental challenge, which we refer to as \textbf{crosstalk} -- unintended information interference across predictive factors~\cite{Shih19}. In cross-stock graph learning, this issue typically arises because models first encode stock sequences into unified embeddings and then propagate them over one or multiple relational graphs. As illustrated in Figure~\ref{fig:framework1}, existing methods follow a two-stage paradigm: unified temporal encoding followed by graph propagation, where information that should remain isolated becomes entangled during cross-stock interaction modeling. We formalize this challenge through two distinct forms of crosstalk.

\textbf{Challenge 1: Temporal-scale Crosstalk.} Financial time series are inherently multi-scale: a stock sequence superposes long-range trends, short-horizon fluctuations, and event-driven shocks~\cite{Chengyunyao23}. These components carry distinct predictive semantics: trends are transferable across related stocks~\cite{Thompson25,SONG2023119236}, while fluctuations capture local market dynamics and shocks are largely stock-specific. However, existing methods encode the entire sequence into a unified representation~\cite{Xuchengfeng22,Lazcano23}, inducing temporal-scale crosstalk that allows asset-specific dynamics to propagate across the stock universe during subsequent graph-based interaction modeling.

\textbf{Challenge 2: Structural Crosstalk.} Stocks are connected through multiple qualitatively different relations (e.g., industry, region, latent co-movement)~\cite{Ekmekcioglu13,Alfaro04,Chen24,Qian24}. Existing methods either construct a single undifferentiated graph or na\"{\i}vely fuse multiple adjacency matrices~\cite{Niu25,Xue25}, obscuring the distinct predictive signal of each individual structure.

Therefore, to systematically mitigate both forms of crosstalk, we propose the \underline{A}nti-\underline{C}ross\underline{T}alk (\textbf{ACT}) framework (Figure~\ref{fig:framework1}) for cross-sectional stock ranking via temporal disentanglement and structural purification. ACT first introduces \textbf{Temporal Component Decomposition (TCD)} to disentangle each stock sequence into trend, fluctuation, and shock components. The fluctuation and shock components then use \textbf{FCI} (temporal convolution for short-horizon oscillations) and \textbf{SCI} (counterfactual buffer for event-driven deviations) to extract component-specific local information and decouple non-transferable dynamics from cross-stock propagation. On the trend branch, \textbf{PSPE} performs progressive structural purification over multi-relational graphs to  isolate relation-specific signals and mitigate structural crosstalk. Finally, \textbf{ACF} adaptively recombines all branch representations for ranking and forecasting. Extensive experiments on two representative datasets demonstrate that ACT significantly outperforms \textbf{16} competitive baselines in both cross-sectional ranking accuracy and portfolio return, achieving improvements of up to \textbf{74.25\%} on the CSI300 dataset. Our main contributions are summarized below.

\begin{itemize}[leftmargin=*]
    \item \textbf{Conceptually:} For the first time, we identify Crosstalk as a key bottleneck in graph-based cross-sectional stock ranking, and systematically characterize it from both temporal-scale and structural perspectives.
    \item \textbf{Methodologically:} We propose the ACT framework, which mitigates two forms of crosstalk through temporal disentanglement (TCD with dedicated FCI/SCI branches) and structural purification (PSPE).
    \item \textbf{Empirically:} Extensive experiments on CSI 300/500 demonstrate that ACT achieves state-of-the-art ranking accuracy and superior portfolio performance, with comprehensive ablation studies further validating the necessity of each anti-crosstalk design.
\end{itemize}

\section{Related Work.}

We review two research lines most relevant to our ACT framework: graph-based cross-sectional stock ranking and time-series decomposition for financial forecasting. Together, these studies motivate our anti-crosstalk perspective on structural and temporal interference.

\subsection{Graph-Based Cross-Sectional Stock Ranking.}
Graph neural networks (GNNs) have become a mainstream paradigm for modeling cross-stock dependence. Early works use static graphs based on predefined relations, such as industry classification~\cite{Thompson25} and supply-chain links~\cite{Liwei21}, with GCN-based propagation~\cite{Kipf17} for representation learning. Subsequent studies further introduce attention~\cite{Velickovic17}, spatio-temporal architectures~\cite{Shao22,Ma23,Wang20,Xu22}, hierarchical aggregation~\cite{Xuchengfeng22,Xue25}, temporal graph modeling~\cite{Wei25,Chen24}, and multi-relational structures~\cite{Qian24,Duan25} to better capture cross-stock dependencies. More recent works also consider richer graph constructions, including anomaly-aware graphs~\cite{Li25}, state-based models~\cite{hu2025finmamba}, hidden-concept discovery in HIST~\cite{xu2022hist}, and historical/future pattern modeling in DishFT-GNN~\cite{Liu_2025_dish}. 

However, these methods either collapse heterogeneous relations into a single graph or na\"{\i}vely fuse multiple adjacency matrices without resolving their mutual interference~\cite{Niu25}. We identify this issue as \textbf{Crosstalk} and propose a principled anti-crosstalk framework.

\subsection{Time-Series Decomposition in Financial Forecasting.}
Time-series decomposition separates an observed sequence into interpretable components. Classical models such as ARIMA~\cite{Shumway17} use moving averages and differencing, and related filters remain common in econometrics~\cite{Babu14}. Recent deep learning models incorporate learnable decomposition into neural architectures~\cite{wu2022autoformer,zhou2022fedformer}, and later studies extend it to guided temporal adaptation~\cite{Chengyunyao23,wu2023timesnet}. In finance, decomposition is also used to extract economically meaningful patterns from market sequences~\cite{griveaubillion2020}.

However, existing decomposition methods mainly aim at improving sequence encoding quality and rarely consider how decomposed components interact with downstream cross-stock propagation. In contrast, our method uses decomposition as an information identification and routing mechanism: rather than only improving component-wise encoding, it separates signals with different transfer characteristics into dedicated pathways to mitigate temporal-scale crosstalk.

\section{Preliminaries.}

\begin{definition}[Multivariate Stock Time Series]
Let $\mathcal{U} = \{u_1, \dots, u_N\}$ denotes the set of $N$ target stocks. At time step $t$, the market state is represented by a feature matrix $\mathbf{X}_t \in \mathbb{R}^{N \times F}$, where $F$ is the number of features. Over a look-back window of length $T$, the historical observation tensor is defined as
\begin{equation}
\mathcal{X} = [\mathbf{X}_{t-T+1}, \dots, \mathbf{X}_t] \in \mathbb{R}^{T \times N \times F}.
\end{equation}
\end{definition}

\begin{definition}[Heterogeneous Stock Graph]
To model cross-stock interactions, we define a heterogeneous graph $\mathcal{G} = (\mathcal{U}, \mathcal{E}, \mathcal{R})$. Its static relation set is $\mathcal{R}_{s} = \{\mathrm{ind}, \mathrm{reg}\}$, corresponding to an industry graph $\mathbf{A}^{\mathrm{ind}} \in \{0,1\}^{N \times N}$ and a region graph $\mathbf{A}^{\mathrm{reg}} \in \{0,1\}^{N \times N}$. These two adjacency matrices encode whether two stocks belong to the same industry sector or the same registration region. In addition, a dynamic adjacency matrix $\mathbf{A}^{d} \in \mathbb{R}^{N \times N}$ is constructed adaptively from projected node representations to capture residual cross-stock dependencies not explained by static relations.
\end{definition}

\begin{definition}[Cross-Sectional Stock Ranking] 
Given $\mathcal{X}$ and $\mathcal{G}$, the goal is to learn a mapping function $f_{\theta}$ that predicts next-step returns and ranks stocks on each trade day. Following standard practice, returns are computed from the volume-weighted average price (VWAP), where $\mathrm{VWAP}_{t} = \sum_{k=1}^{K} p_{k}^{(t)} v_{k}^{(t)} / \sum_{k=1}^{K} v_{k}^{(t)}$. The prediction target and model output are
\begin{equation}
 y_{i,t+1} = \frac{\mathrm{VWAP}_{i,t+1} - \mathrm{VWAP}_{i,t}}{\mathrm{VWAP}_{i,t}}
 \end{equation}
 \begin{equation}
 \hat{\mathbf{Y}}_{t+1} = f_{\theta}(\mathcal{X}, \mathcal{G}) \in \mathbb{R}^{N}
\end{equation}
and the final induced cross-sectional stock ranking is obtained by \begin{equation}
 \pi_{t+1} = \operatorname{Rank}\!\left(\hat{\mathbf{Y}}_{t+1}\right).
\end{equation}
Here, $\pi_{t+1}$ denotes the ranking permutation over all stocks at time $t+1$, which preserves the relative ordering of stocks for downstream portfolio selection.
\end{definition}

\section{The ACT Framework.}

\begin{figure*}[t]
    \centering
    \includegraphics[width=0.6\linewidth]{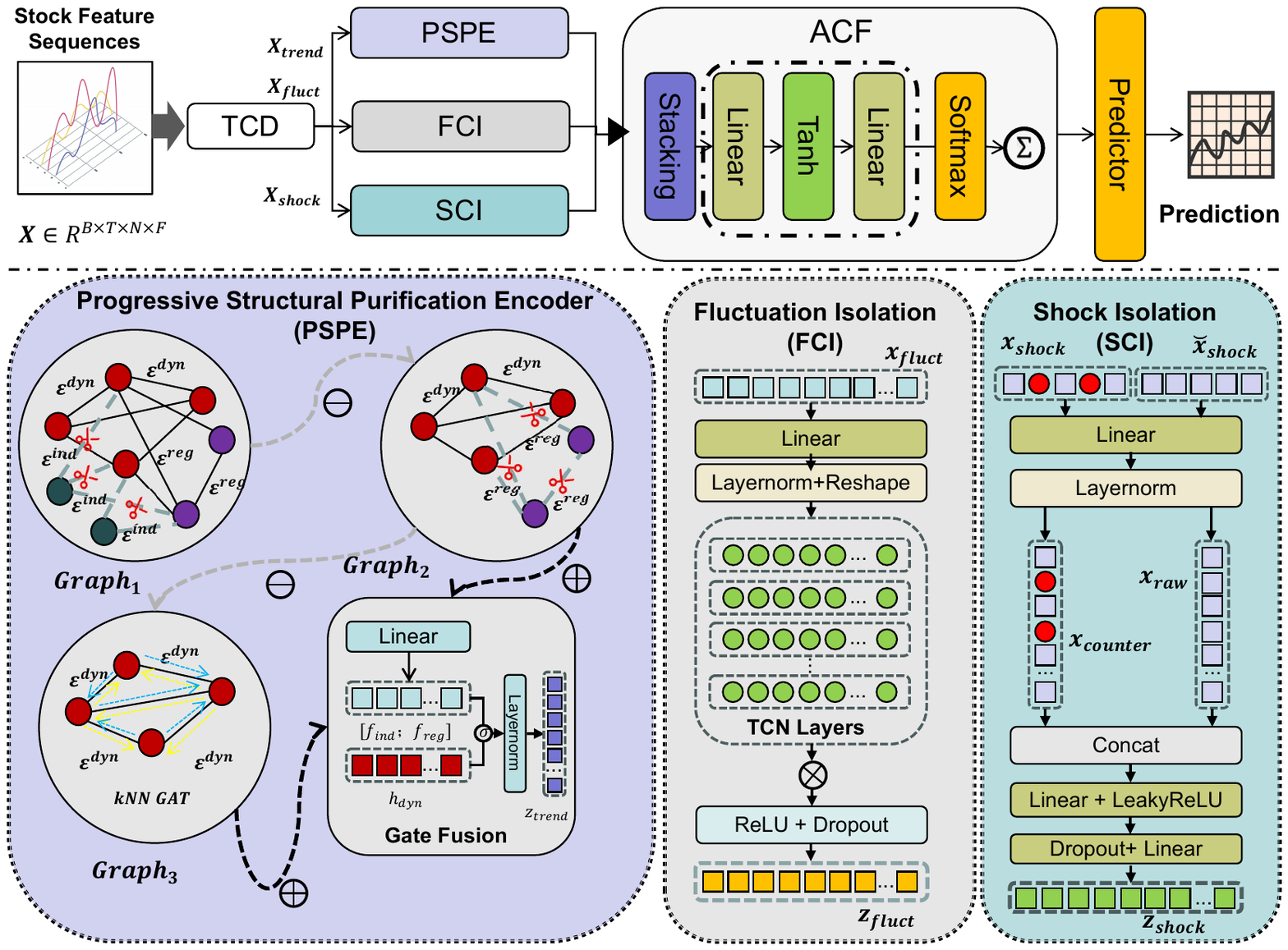}
    \caption{Overall architecture of the ACT framework. The input is decomposed into trend, fluctuation, and shock components via TCD. The trend component is processed by the Progressive Structural Purification Encoder (PSPE), while fluctuation and shock are routed to dedicated isolation branches (FCI and SCI). All representations are recombined by Adaptive Component Fusion (ACF) for final forecasting.}
    \label{fig:framework}
\end{figure*}

As illustrated in Figure~\ref{fig:framework}, ACT processes the input tensor $\mathcal{X} \in \mathbb{R}^{B \times T \times N \times F}$ through five modules. A TCD module first decomposes each stock sequence into trend, fluctuation, and shock components, which are then routed to three dedicated branches: PSPE resolves structural crosstalk in the trend component, FCI isolates short-horizon oscillatory dynamics via temporal convolution, and SCI captures stock-specific event disturbances through a counterfactual buffer. Finally, an Adaptive Component Fusion (ACF) layer dynamically weights the three representations for return forecasting and stock ranking.

\subsection{Temporal Component Decomposition (TCD).}
To mitigate temporal-scale crosstalk at its source, we perform a principled decomposition of the stock sequence $\mathcal{X}$ into trend, fluctuation, and shock components via recursive causal moving averages:
\begin{equation}
\mathbf{X}^{(\cdot)}_{t} = \frac{1}{\min(t{+}1,\, w)}\sum_{k=\max(0,\, t-w+1)}^{t} \mathbf{S}_{k},
\end{equation}
where $(\cdot)=\mathrm{trend}$ with $w=\tau$ and $\mathbf{S}=\mathcal{X}$, while $(\cdot)=\mathrm{fluct}$ with $w=\sigma$ and $\mathbf{S}=\mathbf{X}^{\mathrm{det}}$, where $\mathbf{X}^{\mathrm{det}}_{t} = \mathcal{X}_{t} - \mathbf{X}^{\mathrm{trend}}_{t}$, represents detrend component. The shock component is the residual $\mathbf{X}^{\mathrm{shock}}_{t} = \mathcal{X}_{t} - \mathbf{X}^{\mathrm{trend}}_{t} - \mathbf{X}^{\mathrm{fluct}}_{t}$. Boundary NaN values are set to zero. This decomposition separates heterogeneous temporal semantics before cross-stock interaction is modeled.

\subsection{Progressive Structural Purification Encoder (PSPE).}
PSPE encodes the trend component through three steps: Static Relation Purification, Dynamic Structural Refinement, and Adaptive Fusion. Let $\phi(\cdot)$ denotes LeakyReLU activation function and let $\mathcal{R}_{s} = \{\mathrm{ind}, \mathrm{reg}\}$ denotes the set of static relation types. Note that our method is flexible and can be easily extended to accommodate more relations. 

In the first step, the final trend snapshot is projected into the hidden space:
\begin{equation}
\mathbf{x}^{0} = \operatorname{LN}\!\big(\operatorname{Proj}(\mathbf{X}^{\mathrm{trend}}_{T})\big) 
\end{equation}
and filtered by bidirectional residual purification mechanism:
\begin{equation}
\begin{aligned}
\mathbf{h}_{r} &= \phi\!\big(\operatorname{GCN}_{r}(\mathbf{x}^{0}, \mathbf{A}^{r})\big), \quad r \in \mathcal{R}_{s}\\
\mathbf{f}_{r} &= \phi(W_{r}^{f}\mathbf{h}_{r}), \quad
\mathbf{b}_{r} = W_{r}^{b}\mathbf{h}_{r}, \\
\mathbf{u} &= \mathbf{x}^{0} - \sum_{r \in \mathcal{R}_{s}} \mathbf{b}_{r}, \quad
\mathbf{z}_{s} = W_{s}[\mathbf{f}_{\mathrm{ind}} \| \mathbf{f}_{\mathrm{reg}}].
\end{aligned}
\end{equation}
Here, industry and region graphs are treated as two instances of the same static purification mechanism. For each relation type $r$, PSPE produces a forward output $\mathbf{f}_{r}$ to preserve useful relation-aware signals and a backward output $\mathbf{b}_{r}$ to estimate structural interference; the backward output is subtracted from $\mathbf{x}^{0}$ to form the purified residual $\mathbf{u}$, while $\mathbf{f}_{r}$ is retained as the output in $\mathbf{z}_{s}$ for fusion. This bidirectional residual design reduces interference, as the backward branch models the relation-specific representation under each static graph, and the subtraction operation ensures that signals explained by a given graph (e.g., industry) do not propagate to other graphs (e.g, region) or into subsequent dynamic graph refinement.

In the second step, a dynamic graph supplements residual cross-stock patterns not explained by the static relations. We first project the purified residual as $\tilde{\mathbf{u}} = \phi(W_{d}\mathbf{u})$ and compute pairwise cosine similarities:
\begin{equation}
s_{ij} = \frac{\tilde{\mathbf{u}}_{i}^{\top}\tilde{\mathbf{u}}_{j}}{\|\tilde{\mathbf{u}}_{i}\|_{2}\,\|\tilde{\mathbf{u}}_{j}\|_{2}}.
\end{equation}
A dynamic $k$-NN graph $\mathbf{A}^{d}$ is then constructed by retaining, for each stock $i$, the top-$k$ neighbors with the largest similarity scores:
\begin{equation}
A_{ij}^{d} =
\begin{cases}
1, & j \in \operatorname{TopK}_{k}(s_{i,:}), \\
0, & \text{otherwise},
\end{cases}
\qquad
\mathcal{N}_{d}(i) = \{j \mid A_{ij}^{d} = 1\}.
\end{equation}
On this sparse graph, we use attention-based $\operatorname{GAT}(\cdot, \cdot)$ to pass message on the dynamic $k$-NN graph:
\begin{equation}
\begin{aligned}
e_{ij} &= \phi\!\big(\mathbf{a}^{\top}[\mathbf{W}\tilde{\mathbf{u}}_{i} \| \mathbf{W}\tilde{\mathbf{u}}_{j}]\big), \\
\alpha_{ij} &= \frac{\exp(e_{ij})}{\sum_{j' \in \mathcal{N}_{d}(i)} \exp(e_{ij'})}, \quad j \in \mathcal{N}_{d}(i), \\
\mathbf{H}_{d} &= \operatorname{GAT}(\tilde{\mathbf{u}}, \mathbf{A}^{d})
\end{aligned}
\end{equation}
Here, $\mathbf{a}$ is a learnable attention vector and is distinct from the dynamic adjacency matrix $\mathbf{A}^{d}$. The dynamic representation is then obtained by:
\begin{equation}
(\mathbf{z}_{d})_{i} = \phi\!\left(W_{o} \sum_{j \in \mathcal{N}_{d}(i)} \alpha_{ij} \, \mathbf{W}\tilde{\mathbf{u}}_{j}\right),
\end{equation}
or equivalently $\mathbf{z}_{d} = \phi(W_{o}\mathbf{H}_{d})$.

Finally, static and dynamic representations are adaptively fused to obtain the purified trend embedding:
\begin{equation}
\begin{aligned}
g &= \sigma\!\big(\operatorname{MLP}([\mathbf{z}_{s} \| \mathbf{z}_{d}])\big), \\
\mathbf{z}^{\mathrm{trend}} &= \operatorname{LN}\!\big(\mathbf{z}_{s} + g \odot \mathbf{z}_{d}\big).
\end{aligned}
\end{equation}
This design preserves complementary static priors while allowing dynamic relations to recover residual market structure.

\subsection{Fluctuation Component Isolation (FCI).}
FCI captures short-horizon oscillatory dynamics through a stock-local temporal convolutional pathway, preventing them from leaking into graph propagation. The fluctuation sequence is first projected and reshaped as:
\begin{equation}
\mathbf{F} = \operatorname{Reshape}\!\big(\operatorname{LN}(\operatorname{Proj}(\mathbf{X}^{\mathrm{fluct}}))\big) \in \mathbb{R}^{N \times d \times T}.
\end{equation}
We employ a gated Temporal Convolutional Network (TCN) to capture oscillatory patterns:
\begin{equation}
\begin{aligned}
\mathbf{P} &= \operatorname{Conv}_{1}(\mathbf{F}), \quad
\mathbf{Q} = \sigma\!\big(\operatorname{Conv}_{2}(\mathbf{F})\big), \quad
\mathbf{R} = \operatorname{Conv}_{3}(\mathbf{F}), \\
\mathbf{Z} &= \operatorname{Drop}\!\big(\operatorname{ReLU}(\mathbf{P} \odot \mathbf{Q} + \mathbf{R})\big), \quad
\mathbf{z}^{\mathrm{fluct}} = \mathbf{Z}_{:,:,T}.
\end{aligned}
\end{equation}
Because the convolutions are applied independently to each stock, local oscillations remain isolated from cross-stock message passing.

\subsection{Shock Component Isolation (SCI).}
SCI models stock-specific event disturbances through a compact counterfactual buffer. Since shock signals are often sparse and high-variance, directly encoding them may lead to overfitting to transient outliers. We therefore introduce a smoothed counterfactual sequence that approximates the underlying pattern in the absence of extreme shocks, and learn it jointly with the raw shock sequence. This contrastive design enables the model to better identify and capture extreme event signals through their deviation from the smoothed baseline. Let $\mathbf{S}_{t} = \mathbf{X}^{\mathrm{shock}}_{t}$ denote the shock sequence. We first construct a smoothed counterfactual sequence:
\begin{equation}
\bar{\mathbf{S}}_{t} = \frac{1}{\min(t{+}1,\, w_{s})} \sum_{k=\max(0,\, t-w_{s}+1)}^{t} \mathbf{S}_{k},
\end{equation}
where $w_{s}$ is the smoothing window. The raw and counterfactual embeddings are then obtained by:
\begin{equation}
\begin{aligned}
\mathbf{x} &= \operatorname{LN}\!\big(\operatorname{Proj}(\mathbf{S}_{T})\big), \\
\mathbf{x}' &= \operatorname{LN}\!\big(\operatorname{Proj}(\bar{\mathbf{S}}_{T})\big).
\end{aligned}
\end{equation}
The final shock representation is produced by a two-layer stock-local MLP:
\begin{equation}
\mathbf{z}^{\mathrm{shock}} = W_{2} \, \operatorname{Drop}\!\big(\phi(W_{1}[\mathbf{x} \| \mathbf{x}'])\big).
\end{equation}
This design enables the model to jointly learn each stock's current state and a no-shock counterfactual representation, making the component robust to outlier magnitudes while preventing event noise from contaminating stock graph learning.

\subsection{Adaptive Component Fusion (ACF).}
The three branch representations are stacked as:
\begin{equation}
\mathbf{C} = [\mathbf{z}^{\mathrm{trend}}, \mathbf{z}^{\mathrm{fluct}}, \mathbf{z}^{\mathrm{shock}}] \in \mathbb{R}^{N \times 3 \times d}.
\end{equation}
For stock $i$ and component $c \in \{1,2,3\}$, a two-layer attention MLP computes:
\begin{align}
s_{i,c} &= W_{2}^{a} \, \tanh\!\big(W_{1}^{a}\mathbf{C}_{i,c,:} + b_{1}^{a}\big)
\end{align}
The attention weights are then obtained via softmax function over components, and the final representation is computed as a weighted aggregation:
\begin{equation}
\begin{aligned}
\alpha_{i,c} &= \frac{\exp(s_{i,c})}{\sum_{c'=1}^{3} \exp(s_{i,c'})}, \\
\mathbf{z}_{i} &= \sum_{c=1}^{3} \alpha_{i,c} \, \mathbf{C}_{i,c,:}, \\
\hat{y}_{i} &= W_{\mathrm{out}}\mathbf{z}_{i}
\end{aligned}
\end{equation}
This adaptive fusion allows each stock to weight trend, fluctuation, and shock information differently.

\subsection{Loss Function.}
 For each batch, let $\Omega = \{i \mid y_{i} \text{ is observed}\}$ and define $y_{i}^{c} = \operatorname{clip}(y_{i}, -0.1, 0.1)$. We combine IC loss and MSE loss so that the model learns both correct cross-sectional ranking and reasonable return magnitude calibration. The primary objective is the IC loss:
\begin{equation}
\mathcal{L}_{\mathrm{IC}} = 1 - \frac{\sum_{i \in \Omega}(\hat{y}_{i} - \bar{\hat{y}})(y_{i}^{c} - \bar{y}^{c})}{\sqrt{\sum_{i \in \Omega}(\hat{y}_{i} - \bar{\hat{y}})^{2} + \epsilon} \, \sqrt{\sum_{i \in \Omega}(y_{i}^{c} - \bar{y}^{c})^{2} + \epsilon}},
\end{equation}
where $\bar{\hat{y}}$ and $\bar{y}^{c}$ are means over $\Omega$. The auxiliary loss is:
\begin{equation}
\mathcal{L}_{\mathrm{MSE}} = \frac{1}{|\Omega|} \sum_{i \in \Omega}(\hat{y}_{i} - y_{i}^{c})^{2}
\end{equation}
which provides auxiliary supervision on return magnitude and complements the IC-based objective, thereby helping further improve cross-sectional ranking performance. The final objective function is defined as:
\begin{equation}
\mathcal{L} = \mathcal{L}_{\mathrm{IC}} + \lambda \mathcal{L}_{\mathrm{MSE}}, \quad \lambda \in [0,1]
\end{equation}

\subsection{Inference.} During inference, a sliding window of length $T$ is applied over the test horizon. For each window, ACT outputs a score vector $\hat{\mathbf{y}} \in \mathbb{R}^{N}$ for the final timestamp within the window. The resulting daily scores are then flattened into a prediction series indexed by $(\mathrm{datetime}, \mathrm{instrument})$ for downstream cross-sectional stock ranking evaluation.

\section{Experiments.}

\begin{table}[h]
\centering
\caption{Dataset Details}
\small
\renewcommand{\arraystretch}{0.8}
\resizebox{\linewidth}{!}{
\begin{tabular}{lcc}
\toprule
 & CSI300 & CSI500 \\
\midrule
Stock number & 300 & 500 \\
Training period & 04/01/2010--31/12/2020 & 04/01/2010--31/12/2020 \\
Validation period & 01/01/2021--31/12/2022 & 01/01/2021--31/12/2022 \\
Testing period & 01/01/2023--30/09/2025 & 01/01/2023--30/09/2025 \\
\bottomrule
\end{tabular}
}
\label{tab:datasets}
\end{table}

\begin{table*}[t]
\centering
\caption{Performance Comparison on CSI300 and CSI500 Datasets, with the best value \textbf{bold} and the suboptimal value \underline{underlined}. Improvement (\%) is relative to the suboptimal baseline.}
\small
\renewcommand{\arraystretch}{0.9}
\resizebox{\textwidth}{!}{
\begin{tabular}{lcccccccc}
\toprule
\multirow{2}{*}{Model} & \multicolumn{4}{c}{CSI300} & \multicolumn{4}{c}{CSI500} \\
\cmidrule(lr){2-5} \cmidrule(lr){6-9}
 & IC$\uparrow$ & ICIR$\uparrow$ & RankIC$\uparrow$ & RankICIR$\uparrow$ & IC$\uparrow$ & ICIR$\uparrow$ & RankIC$\uparrow$ & RankICIR$\uparrow$ \\
\midrule
LSTM & 0.0276 & 0.2239 & 0.0347 & 0.3312 & 0.0466 & 0.4623 & 0.0522 & 0.5500 \\
GRU & 0.0310 & 0.2708 & 0.0399 & 0.3756 & 0.0578 & 0.5608 & 0.0646 & \underline{0.6971} \\
Transformer & 0.0329 & 0.2809 & 0.0394 & 0.3650 & 0.0529 & 0.4706 & 0.0523 & 0.5693 \\
ALSTM & 0.0355 & 0.3108 & 0.0395 & 0.3712 & 0.0506 & 0.4625 & 0.0580 & 0.5784 \\
SFM & 0.0340 & 0.2959 & 0.0306 & 0.3009 & 0.0274 & 0.2449 & 0.0304 & 0.3362 \\
GAT & 0.0346 & 0.2609 & 0.0216 & 0.1835 & 0.0450 & 0.3195 & 0.0382 & 0.3045 \\
TCN & 0.0349 & 0.3109 & 0.0410 & 0.3739 & 0.0593 & 0.5971 & 0.0523 & 0.5653 \\
TabNet & 0.0294 & 0.2478 & 0.0326 & 0.3089 & 0.0320 & 0.3069 & 0.0331 & 0.3620 \\
Localformer & 0.0285 & 0.2439 & 0.0399 & 0.3728 & 0.0466 & 0.4275 & 0.0554 & 0.5901 \\
\cmidrule(lr){1-9}
XGBoost & \underline{0.0528} & \underline{0.4338} & \underline{0.0476} & \underline{0.4482} & 0.0601 & 0.5484 & 0.0540 & 0.6102 \\
CatBoost & 0.0350 & 0.2991 & 0.0360 & 0.3491 & 0.0449 & 0.3707 & 0.0410 & 0.4336 \\
LightGBM & 0.0360 & 0.3024 & 0.0371 & 0.3453 & 0.0442 & 0.3604 & 0.0409 & 0.4266 \\
DoubleEnsemble & 0.0270 & 0.2239 & 0.0332 & 0.3117 & 0.0331 & 0.2802 & 0.0340 & 0.3343 \\
\cmidrule(lr){1-9}
iTransformer   & 0.0456 & 0.3390 & 0.0301 & 0.2508 & \underline{0.0845} & \underline{0.6325} & \underline{0.0707} & 0.5858 \\
TimeMixer      & 0.0234 & 0.1899 & 0.0137 & 0.1135 & 0.0349 & 0.2670 & 0.0384 & 0.3133 \\
FreqCycle      & 0.0325 & 0.2519 & 0.0253 & 0.2136 & 0.0348 & 0.2634 & 0.0344 & 0.2664 \\
\cmidrule(lr){1-9}
\textbf{ACT} & \textbf{0.0692} & \textbf{0.6955} & \textbf{0.0786} & \textbf{0.7810} & \textbf{0.0877} & \textbf{0.8522} & \textbf{0.0907} & \textbf{0.9216} \\
\midrule
Improvement (\%) & 31.06 & 60.33 & 65.13 & 74.25 & 3.79 & 34.74 & 28.29 & 32.20 \\
\bottomrule
\end{tabular}
}
\label{tab:rq1}
\end{table*}

In this section, we evaluate ACT from five perspectives: overall comparison with representative baselines (\hyperref[section5.2]{section 5.2}), contribution of the anti-crosstalk modules (\hyperref[section5.3]{section 5.3}), portfolio backtesting performance (\hyperref[section5.4]{section 5.4}), alpha robustness and factor mining (\hyperref[section5.5]{section 5.5}), and subgroup analysis for validity of relation structure (\hyperref[subgroup_analysis]{Appendix A.3}).

\subsection{Experiment Setting.}
We summarize the experimental protocol, including the datasets, compared baselines, and evaluation metrics used throughout all experiments in this section.

\subsubsection{Dataset.}
We evaluate our ACT framework on two publicly available datasets: CSI300 and CSI500, which represent major Chinese stock market. Besides, We use the stock features of Alpha158 in the open-source quantitative investment platform Qlib. For each stock on date, Alpha158 looks back 40 days as length of stock sequence. The datasets are split into training (04/01/2010--31/12/2020), validation (01/01/2021--31/12/2022), and testing (01/01/2023--30/09/2025) periods as detailed in Table~\ref{tab:datasets}. The industry relation settings follow the CSRC (China Securities Regulatory Commission) rule while region relations are categorized by province granularity, both from CSMAR.\footnote{China Stock Market \& Accounting Research Database (\url{https://data.csmar.com/}).}

\subsubsection{Compared Methods.}
To comprehensively evaluate ACT, we select a diverse set of representative deep
learning baselines that have achieved remarkable results in the fields of quantitative finance and time series forecasting: LSTM~\cite{Hochreiter97}, GRU~\cite{cho2014gru}, Transformer~\cite{vaswani2023attention}, ALSTM~\cite{qin2017alstm}, SFM~\cite{sfm}, GAT~\cite{Velickovic17}, Localformer (original from Qlib), TCN~\cite{tcn}, TabNet~\cite{tabnet}, XGBoost~\cite{chen2016xgboost}, CatBoost~\cite{catboost}, LightGBM~\cite{ke17}, DoubleEnsemble~\cite{Chuhengzhang20}, iTransformer~\cite{liu2023itransformer}, TimeMixer~\cite{wang2023timemixer}, FreqCycle~\cite{zhang2026freqcycle}. These diverse baselines enable a systematic comparison from the perspectives of temporal and interaction modeling.

\subsubsection{Evaluation Metrics.}
We evaluate all methods using four widely adopted evaluation metrics in stock ranking area: Information Coefficient (IC), IC Information Ratio (ICIR), Rank IC, and Rank ICIR, which measure predictive quality in terms of linear correlation, stability, and ranking effectiveness.

\subsection{Comparison with Baselines.}
\label{section5.2}

As shown in Table~\ref{tab:rq1}, ACT consistently outperforms all existing methods across both datasets and all metrics, achieving best results. This indicates that the advantage of ACT is not confined to a particular evaluation dimension, but is reflected jointly in predictive accuracy and prediction stability. Besides, ACT demonstrates substantial gains relative to the second-best method per metric, up to 74.25\% on CSI300 and 34.74\% on CSI500. 

\subsection{Ablation Study.}
\label{section5.3}
To systematically evaluate the contribution of each anti-crosstalk module in ACT, we conduct ablation experiments on both CSI300 and CSI500 datasets by removing one key component at a time and replacing it with a simpler counterpart. Specifically, \textbf{w/o PSPE} replaces the Progressive Structural Purification Encoder with a standard GAT, removing the sequential extraction and residual subtraction mechanism; \textbf{w/o FCI} replaces the Fluctuation Component Isolation branch (TCN) with a one-layer MLP, removing the dedicated temporal pathway for oscillatory dynamics; and \textbf{w/o SCI} replaces the Shock Component Isolation branch (Counterfactual Buffer) with a one-layer MLP, removing the counterfactual smoothing and contrastive design. The results are shown in Table~\ref{tab:ablation}.

In summary, the ablation results demonstrate the effectiveness and necessity of each anti-crosstalk module in ACT. Specifically:
\FloatBarrier
\textbf{w/o PSPE (Progressive Structural Purification Encoder)} demonstrates the importance of resolving structural crosstalk. Removing PSPE (replaced by standard GAT) leads to a 25.87\% drop in IC on CSI300 and an 43.67\% drop on CSI500, indicating that the bidirectional residual design is crucial for decoupling heterogeneous relational effects.

\textbf{w/o FCI (Fluctuation Component Isolation)} shows the value of isolating short-horizon dynamics from graph propagation. Compared with the 3.88\% drop observed on the CSI500 dataset, drop on CSI300 reaches 25.00\%, showing that temporal-scale crosstalk from fluctuations plays a stronger role in markets with sparser relational structures.

\textbf{w/o SCI (Shock Component Isolation)} shows a substantial impact across both datasets. Its removal leads to a 44.51\% IC drop on CSI300 and a 25.88\% drop on CSI500, demonstrating that isolating idiosyncratic event-driven disturbances is vital for preventing temporal-scale crosstalk. This confirms that stock-specific shocks can dominate short-term returns in certain circumstances.

\begin{table}[t]
\caption{Ablation study on CSI300/CSI500. Each row reports absolute performance, with following row reporting relative change: $\Delta=\frac{\text{Variant}-\text{ACT}}{\text{ACT}}\times 100\%$.}
\label{tab:ablation}
\centering
\renewcommand{\arraystretch}{0.7}
\footnotesize
\setlength{\tabcolsep}{8pt}
\resizebox{\columnwidth}{!}{
\begin{tabular}{lcccc}
\toprule
\multicolumn{5}{c}{CSI300} \\
\midrule
Variant & IC & ICIR & RankIC & RankICIR \\
\midrule
w/o PSPE & 0.0513 & 0.4998 & 0.0614 & 0.6018 \\
         & -25.87\% & -28.14\% & -21.88\% & -22.94\% \\
\cmidrule(lr){1-5}
w/o FCI  & 0.0519 & 0.4930 & 0.0600 & 0.5923 \\
         & -25.00\% & -29.11\% & -23.66\% & -24.16\% \\
\cmidrule(lr){1-5}
w/o SCI  & 0.0384 & 0.3744 & 0.0515 & 0.5048 \\
         & -44.51\% & -46.17\% & -34.48\% & -35.36\% \\
\midrule
\textbf{ACT} & \textbf{0.0692} & \textbf{0.6955} & \textbf{0.0786} & \textbf{0.7810} \\
\midrule
\multicolumn{5}{c}{CSI500} \\
\midrule
Variant & IC & ICIR & RankIC & RankICIR \\
\midrule
w/o PSPE & 0.0494 & 0.4994 & 0.0575 & 0.6318 \\
         & -43.67\% & -41.40\% & -36.60\% & -31.44\% \\
\cmidrule(lr){1-5}
w/o FCI  & 0.0843 & 0.8274 & 0.0885 & 0.8908 \\
         & -3.88\% & -2.91\% & -2.43\% & -3.34\% \\
\cmidrule(lr){1-5}
w/o SCI  & 0.0650 & 0.6697 & 0.0738 & 0.7807 \\
         & -25.88\% & -21.41\% & -18.63\% & -15.29\% \\
\midrule
\textbf{ACT} & \textbf{0.0877} & \textbf{0.8522} & \textbf{0.0907} & \textbf{0.9216} \\
\bottomrule
\end{tabular}}
\end{table}

\subsection{Backtesting Performance in Portfolio.}
\label{section5.4}
In this section, we evaluate how ACT's predictive advantage translates into portfolio profitability. All models are backtested using Qlib's spot strategy (TopKDropoutStrategy in K=50 and N=5) on the CSI500 dataset from 2023 to 2025. This is a long-only strategy which selects $K$ best stocks and then discards $N$ stocks. All reported metrics are computed based on excess returns. We adopt the following widely used backtesting metrics, including Annualized Return (AR), Information Ratio (IR), Maximum Drawdown (MD), Cumulative Return (CR), Sharpe Ratio (Sharpe), and Calmar Ratio (Calmar). 

\begin{table}[t]
\centering
\caption{Portfolio backtesting on CSI500 (TopKDropout, K=50, N=5). All metrics except CR are computed on excess returns (without cost),  with the best value \textbf{bold} and the 2nd/3rd value \underline{underlined}.}
\label{tab:rq2_backtest}
\renewcommand{\arraystretch}{1.1}
\setlength{\tabcolsep}{6pt}
\footnotesize
\resizebox{\columnwidth}{!}{
\begin{tabular}{lcccccc}
\toprule
\textbf{Model} & \textbf{AR}$\uparrow$ & \textbf{IR}$\uparrow$ & \textbf{MD}$\downarrow$ & \textbf{CR}$\uparrow$ & \textbf{Sharpe}$\uparrow$ & \textbf{Calmar}$\uparrow$ \\
\midrule
LSTM           & 0.1561 & 0.9069 & $-$0.3267 & 0.8921 & 0.9332 & 0.4779 \\
GRU            & 0.2506 & 1.4325 & $-$0.2933 & 1.4627 & 1.4740 & 0.8544 \\
Transformer    & \underline{0.2934} & \underline{1.6632} & $-$0.2889 & \underline{1.7757} & \underline{1.7114} & \underline{1.0156} \\
ALSTM          & 0.2894 & 1.6316 & $-$0.2973 & 1.7426 & 1.6789 & 0.9734 \\
SFM            & 0.2090 & 1.2045 & $-$0.2833 & 1.1915 & 1.2394 & 0.7377 \\
GAT            & 0.1612 & 0.9130 & $-$0.2964 & 0.9175 & 0.9366 & 0.5439 \\
TCN            & 0.2140 & 1.2193 & $-$0.2879 & 1.2228 & 1.2547 & 0.7434 \\
TabNet         & 0.2290 & 1.2950 & $-$0.2898 & 1.3185 & 1.3325 & 0.7902 \\
Localformer    & 0.2852 & 1.6028 & $-$0.2872 & 1.7096 & 1.6492 & 0.9930 \\
\cmidrule(lr){1-7}
XGBoost        & 0.2720 & 1.5471 & $-$0.3133 & 1.6132 & 1.5919 & 0.8683 \\
CatBoost       & 0.2121 & 1.2025 & $-$0.2926 & 1.2088 & 1.2374 & 0.7251 \\
LightGBM       & 0.2148 & 1.2090 & $-$0.2773 & 1.2239 & 1.2441 & 0.7747 \\
DoubleEnsemble & 0.2390 & 1.3539 &  \textbf{$-$0.2643} & 1.3816 & 1.3931 & 0.9044 \\
\cmidrule(lr){1-7}
iTransformer   & \underline{0.3252} & \underline{1.8040} & $-$0.2749 & \underline{2.0527} & \underline{1.8563} & \underline{1.1829} \\
TimeMixer      & 0.2692 & 1.5015 & $-$0.2790 & 1.6117 & 1.5450 & 0.9650 \\
FreqCycle      & 0.2192 & 1.2600 & \underline{$-$0.2718} & 1.2731 & 1.2965 & 0.8064 \\
\midrule 
 \textbf{ACT} &  \textbf{0.4579} &  \textbf{2.5944} & \underline{$-$0.2733} &  \textbf{3.3955} &  \textbf{2.6696} &  \textbf{1.6754} \\
\bottomrule
\end{tabular}
}
\end{table}

As shown in Table~\ref{tab:rq2_backtest}, ACT achieves dominant backtesting performance on CSI500 across nearly all metrics. ACT's AR of 0.4579 exceeds the second-best iTransformer (0.3252) by 40.1\%, and its IR of 2.5944 surpasses iTransformer's 1.8040 by 43.8\%. The Calmar ratio of 1.6754 indicates that ACT generates significantly more return per unit of tail risk. The cumulative return (CR) of 3.3955 also exceeds the runner-up (iTransformer: 2.0527), while the Sharpe ratio of 2.6696 confirms robust risk-adjusted profitability; note that it is numerically close to IR (2.5944) because both are derived from excess returns under a zero risk-free rate.

Regarding maximum drawdown, ACT's MD ($-$0.2733) ranks competitively but does not achieve the best value (DoubleEnsemble: $-$0.2643), which is expected: ACT's aggressive return-seeking entails higher peak-to-trough drawdowns, but the calmar ratio indicates superior risk-adjusted performance.

\begin{figure}[t]
    \centering
    \includegraphics[width=0.90\columnwidth]{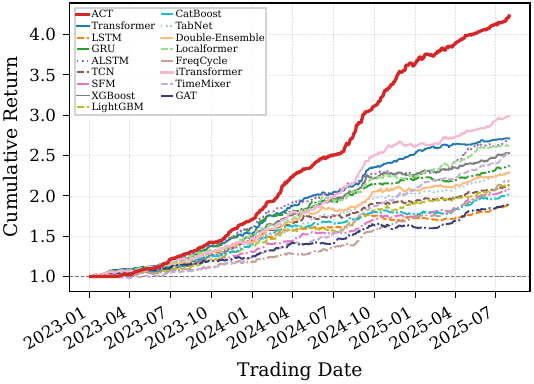}
    \caption{Cumulative Return on CSI500 from 2023 to 2025. ACT (red) substantially outperforms all baselines.}
    \label{fig:cum_return_csi500_1}
\end{figure}

Figure~\ref{fig:cum_return_csi500_1} visualize the cumulative return curves on the CSI500 dataset. The results show that ACT exhibits a pronounced advantage in alpha extraction. The performance gap between ACT and baseline methods is more pronounced than in the overall comparison above: The curve of ACT keeps rising, while other models tend to stabilize during fluctuation periods (such as the middle of 2024).

\subsection{Alpha Robustness and Factor Mining.}
\label{section5.5}
While sections 5.2--5.4 demonstrate ACT's aggregate superiority, a natural concern is whether its gains reflect risk factor exposure or genuine crosstalk mitigation. Therefore, We conduct the Fama-French factor (three-factor and five-factor) regression on the CSI500 dataset.

The core purpose of the regression is to test whether the excess return alpha of an investment strategy can be explained by the representative factors (market, size, value, profitability, investment). 

\subsubsection{Fama-French Regression Settings.}
We construct the excess return series for each model's portfolio as $R_{p,t} - R_{f,t}$, where $R_{p,t}$ is the daily portfolio return from Qlib's TopKDropout backtesting (K=50, N=5) and $R_{f,t}$ is the daily risk-free rate. Both Fama-French factor data (MktRF, SMB, HML, RMW, CMA) and the risk-free rate are obtained from the CSMAR database, covering the test period from January 2023 to September 2025. We estimate both the three-factor model (MktRF, SMB, HML) and the five-factor model (MktRF, SMB, HML, RMW, CMA):
\begin{equation}
R_{p,t} - R_{f,t} = \hat{\alpha} + \sum_{k} \beta_k F_{k,t} + \varepsilon_t
\label{eq:ff_regression}
\end{equation}
where $t$-statistics are computed using Newey-West HAC standard errors with 5 lags to account for serial correlation.

We focus factor regression on ACT and two baseline models that suboptimal on CSI300 and CSI500 datasets respectively (iTransformer and XGBoost). Section~\ref{tab:ff5_csi500} and Tables~\ref{tab:ff3_csi500}(in Appendix) report regression results.

\subsubsection{Analysis.}
Two key findings can be drawn from Tables~\ref{tab:ff5_csi500} and Appendix~\ref{tab:ff3_csi500}. Firstly, all three models produce daily alphas significant at the 1\% level, and ACT's alpha ($\alpha=0.0023$, $t>13$) is greater than that of iTransformer ($0.0007$) and XGBoost ($0.0014$), confirming that ACT captures predictive signals beyond well-known factors.

\begin{table}[!t]
\centering
\caption{Fama--French Five-Factor Regression Results of Daily Return Spreads.}
\label{tab:ff5_csi500}
\renewcommand{\arraystretch}{1.5}
\setlength{\tabcolsep}{2.5pt}
\footnotesize
\resizebox{\columnwidth}{!}{
\begin{tabular}{lccccccccc}
\toprule
\textbf{Model} & $\alpha$ & $t(\alpha)$ & $\beta_m$ & $\beta_s$ & $\beta_h$ & $\beta_r$ & $\beta_c$ & $R^2$ & \textbf{Obs.} \\
\midrule
ACT & 0.0023*** & 13.342 & -0.0011 & -0.0132 & -0.0077 & -0.0195 & 0.0075 & 0.002 & 625 \\
 & (0.0002) &  & (0.011) & (0.017) & (0.030) & (0.036) & (0.038) & & \\
\addlinespace[5pt]
iTransformer & 0.0007*** & 5.096 & 0.0095 & -0.0167 & 0.0168 & 0.0087 & 0.0464 & 0.005 & 664 \\
 & (0.0001) &  & (0.018) & (0.017) & (0.029) & (0.035) & (0.040) & & \\
\addlinespace[5pt]
XGBoost & 0.0014*** & 8.685 & -0.0205 & 0.0128 & 0.0473 & 0.0494 & 0.0153 & 0.008 & 664 \\
 & (0.0002) &  & (0.014) & (0.019) & (0.027) & (0.035) & (0.042) & & \\
\bottomrule
\end{tabular}}

\vspace{2pt}
\parbox{\columnwidth}{\footnotesize\raggedright \textit{Notes.} The dependent variable is the daily portfolio excess return $R_{p,t} - R_{f,t}$. Newey--West (HAC) standard errors with 5 lags are reported in parentheses. *, **, and *** denote statistical significance at the 10\%, 5\%, and 1\% levels, respectively.}
\end{table}

Secondly, ACT's alpha is virtually identical under FF3 and FF5 ($0.0023$ in both), indicating that the profitability (RMW) and investment (CMA) factors add no explanatory power; by contrast, iTransformer shows a non-negligible CMA loading ($\beta_c \approx 0.046$) and XGBoost shows non-negligible HML and RMW loadings ($\beta_h \approx 0.047$, $\beta_r \approx 0.049$) under F-F5.

\section{Conclusion.}
In this study, we identify crosstalk, namely unintended information interference across predictive factors, as a key challenge in cross-sectional stock ranking, and propose a anti-crosstalk framework \textbf{ACT}. ACT mitigates both temporal-scale and structural crosstalk through our temporal disentanglement and structural purification modules, enabling modeling of more transferable cross-stock dependencies. Extensive experiments on two representative datasets demonstrate that ACT consistently outperforms strong baselines in both ranking accuracy and portfolio return.

\FloatBarrier
\bibliographystyle{siamplain}
\bibliography{par/reference}

\appendix
\section{Appendix}
\label{sec:appendix}

\subsection{Experimental Metrics.}

\begin{itemize}
\item \textbf{Annualized Return (AR$\uparrow$):} Annualized excess return of the portfolio over the benchmark.
\item \textbf{Information Ratio (IR$\uparrow$):} Risk-adjusted excess return, defined as $\text{IR} = \frac{\mathbb{E}[r_{\text{excess}}]}{\sigma(r_{\text{excess}})} \times \sqrt{252}$, where $r_{\text{excess}}$ denotes the daily excess return.
\item \textbf{Maximum Drawdown (MD$\downarrow$):} The largest peak-to-trough decline in excess cumulative return.
\item \textbf{Cumulative Return (CR$\uparrow$):} Total absolute portfolio return over the entire testing period.
\item \textbf{Sharpe Ratio (Sharpe$\uparrow$):} Annualized Sharpe ratio computed on excess returns. Note that in this setting, the Sharpe ratio is numerically close to IR because both are computed using excess returns with a zero risk-free rate assumption.
\item \textbf{Calmar Ratio (Calmar$\uparrow$):} Ratio of annualized return to maximum drawdown magnitude, $\text{Calmar} = \text{AR} / |\text{MD}|$, measuring return per unit of tail risk.
\end{itemize}

\subsection{Fama–French Three-Factor Regression Results.}
The following is the table of Fama--French Three-Factor Regression of Daily Return Spreads.

\begin{table}[h]
\centering
\caption{Fama--French Three-Factor Regression of Daily Return Spreads.}
\label{tab:ff3_csi500}
\renewcommand{\arraystretch}{1.35}
\setlength{\tabcolsep}{5pt}
\footnotesize
\resizebox{\columnwidth}{!}{
\begin{tabular}{lccccccc}
\toprule
\textbf{Model} & $\alpha$ & $t(\alpha)$ & $\beta_m$ & $\beta_s$ & $\beta_h$ & $R^2$ & \textbf{Obs.} \\
\midrule
ACT & 0.0023*** & 13.287 & -0.0029 & -0.0076 & 0.0019 & 0.001 & 625 \\
 & (0.0002) &  & (0.011) & (0.013) & (0.023) & & \\
\addlinespace[4pt]
iTransformer & 0.0007*** & 5.040 & 0.0114 & -0.0053 & 0.0217 & 0.003 & 664 \\
 & (0.0001) &  & (0.017) & (0.012) & (0.026) & & \\
\addlinespace[4pt]
XGBoost & 0.0014*** & 8.780 & -0.0152 & 0.0080 & 0.0291 & 0.005 & 664 \\
 & (0.0002) &  & (0.013) & (0.016) & (0.023) & & \\
\bottomrule
\end{tabular}}
\vspace{2pt}
\parbox{\columnwidth}{\footnotesize\raggedright \textit{Notes.} The dependent variable is the daily portfolio excess return $R_{p,t} - R_{f,t}$. Newey--West (HAC) standard errors with 5 lags are reported in parentheses. *, **, and *** denote statistical significance at the 10\%, 5\%, and 1\% levels, respectively.}
\end{table}

\subsection{Subgroup Heterogeneity Analysis.}
\label{subgroup_analysis}
To verify that the inter-stock relationship truly contains valid information and causes crosstalk, we should observe a testable prediction: the gain of ACT should be positively correlated with the local density of the relationship graph. Conversely, if the advantage of ACT comes from factors unrelated to the relationship (such as model capacity or feature engineering), then the gain distribution should be independent of the graph density.

\subsubsection{Subgroup Analysis Settings.}
We partition the 500 stocks in the CSI500 pool into subgroups by industry sector and company registration region. Industry classifications follow the CSRC (China Securities Regulatory Commission) rule, supplemented by data fetched via the AKShare database to fill missing entries to ensure that each stock has its own category(15 sectors). Company regions are determined by company registration province, mapped to 7 macro-regions (North China, East China, South China, Central China, NE China, NW China, SW China) to prevent over-detailed divisions from insufficient sample size in subgroup. For each subgroup, we compute IC, ICIR, Rank IC, and Rank ICIR using the per-stock daily predictions and real returns over test period. This design directly probes whether ACT's advantage concentrates in sectors with dense relational structure---where its graph-based anti-crosstalk mechanism should be most effective---or generalizes broadly.

Note that iTransformer covers fewer stocks than ACT and XGBoost due to stricter input completeness requirements during inference; subgroups where iTransformer retains fewer than 5 stocks are marked ``--'' in tables and omitted from bar charts.

\begin{figure}[t]
    \centering
    \begin{subfigure}[b]{0.24\textwidth}
        \centering
        \includegraphics[width=\textwidth]{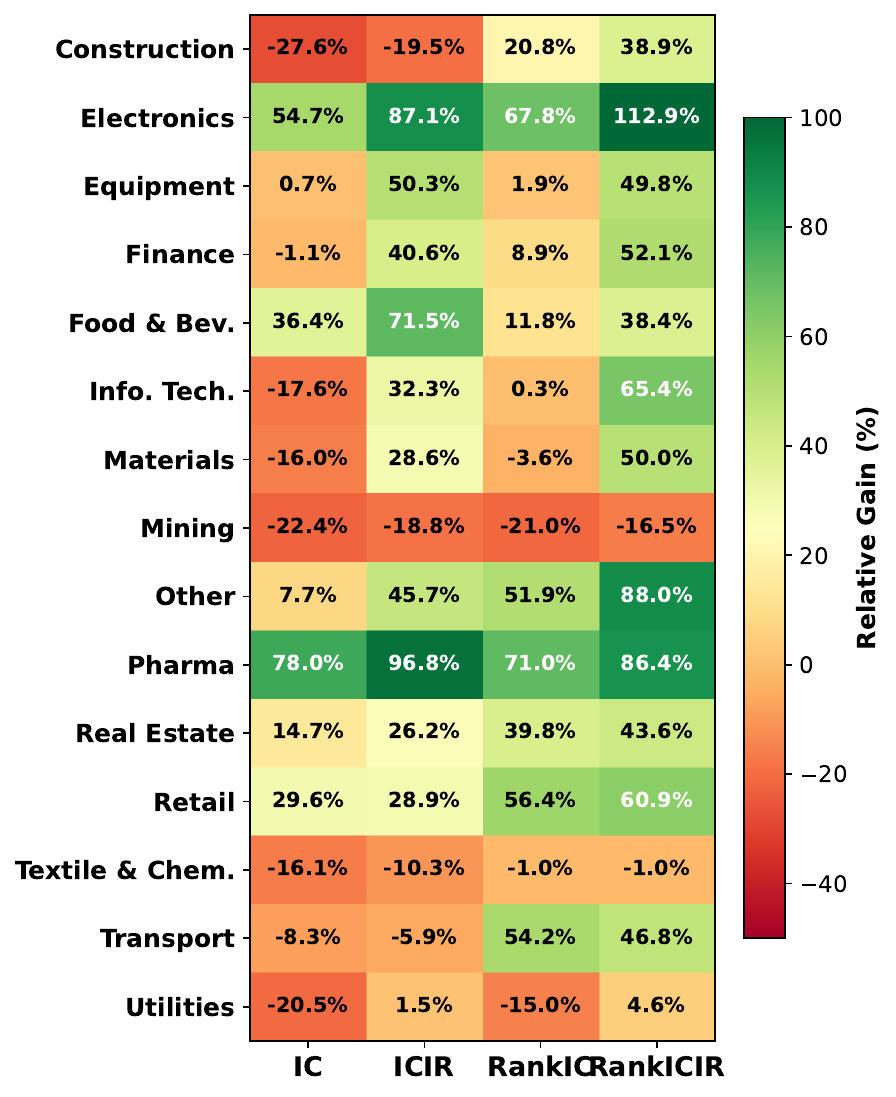}
        \caption{Industry Heatmap}
        \label{fig:industry_heatmap}
    \end{subfigure}%
    \hfill
    \vspace{3pt}
    \begin{subfigure}[b]{0.24\textwidth}
        \centering
        \includegraphics[width=\textwidth]{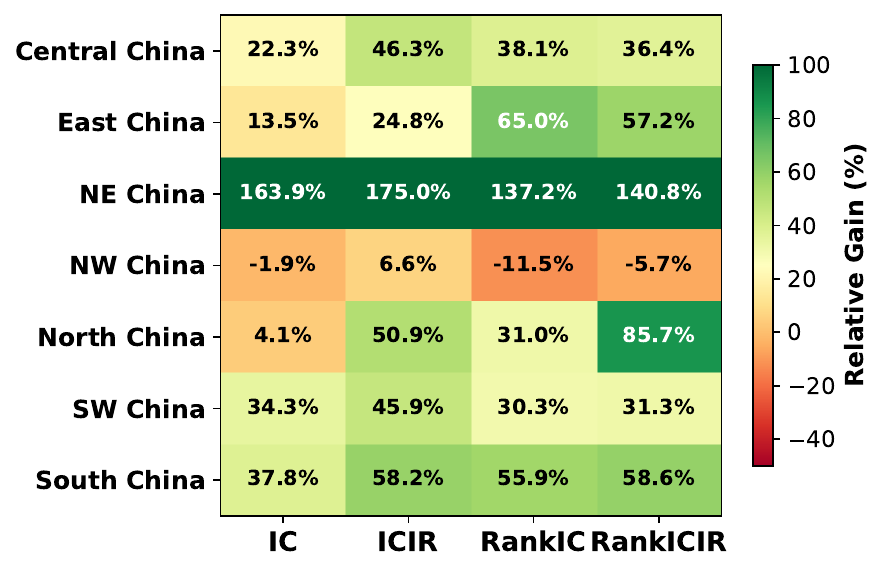}
        \caption{Region Heatmap}
        \label{fig:region_heatmap}
    \end{subfigure}
    \caption{Relative improvement (\%) of ACT over the best performance among two baselines (XGBoost, iTransformer), by industry sector and geographic region on CSI500. Darker green indicates larger gains; red indicates the baseline outperforms ACT.}
    \label{fig:subgroup_heatmaps}
\end{figure}

\subsubsection{Analysis.}
\begin{figure*}[htbp]
    \centering
    \begin{subfigure}[b]{0.95\textwidth}
        \centering
        \includegraphics[width=\textwidth]{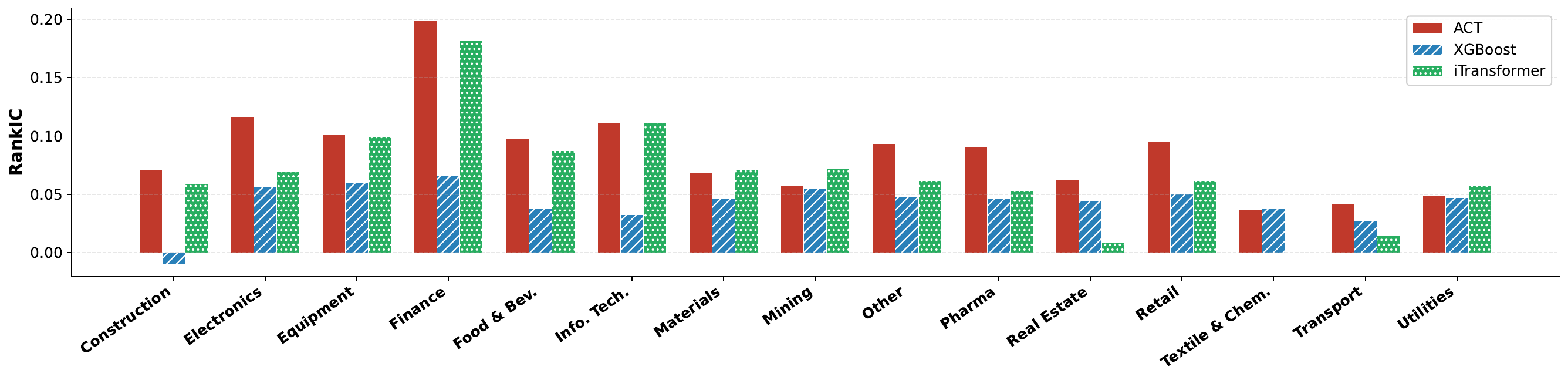}
        \caption{Rank IC by Industry Sector (CSI500)}
        \label{fig:appendix_bar_industry_rankic500}
    \end{subfigure}%
    \vspace{8pt}
    \begin{subfigure}[b]{0.95\textwidth}
        \centering
        \includegraphics[width=\textwidth]{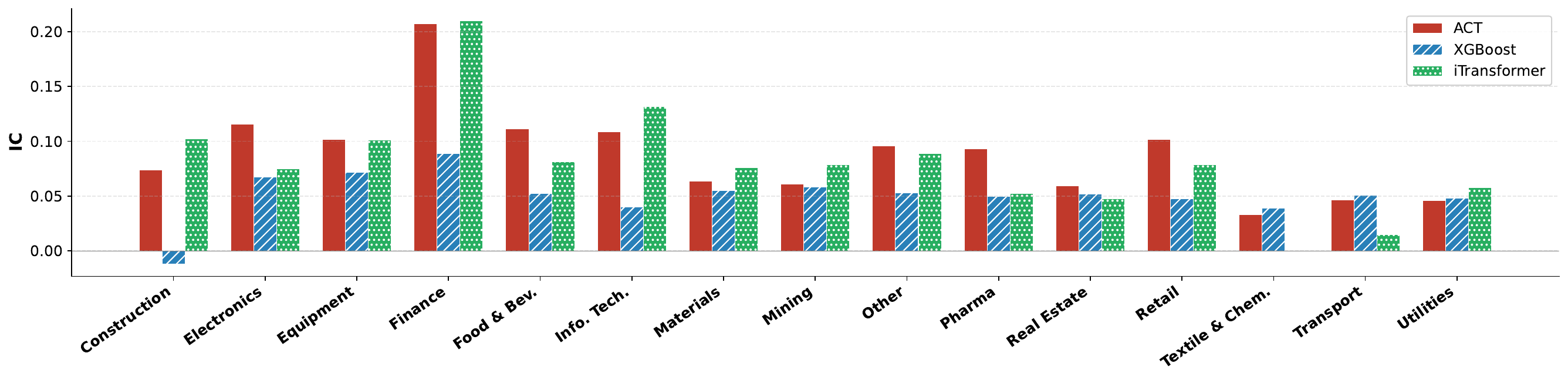}
        \caption{IC by Industry Sector (CSI500)}
        \label{fig:appendix_bar_industry_ic500}
    \end{subfigure}
    \caption{Absolute IC and Rank IC comparison by industry sector on CSI500. ACT (red) dominates in relationally dense sectors (Electronics, Pharma, Food \& Bev.), while iTransformer is competitive in some sectors (Mining, Materials). Companion to the relative-improvement heatmap in Figure~\ref{fig:industry_heatmap}.}
    \label{fig:appendix_bar_industry}
\end{figure*}

\begin{figure*}[h]
    \centering
    \begin{subfigure}[b]{0.48\textwidth}
        \centering
        \includegraphics[width=\textwidth]{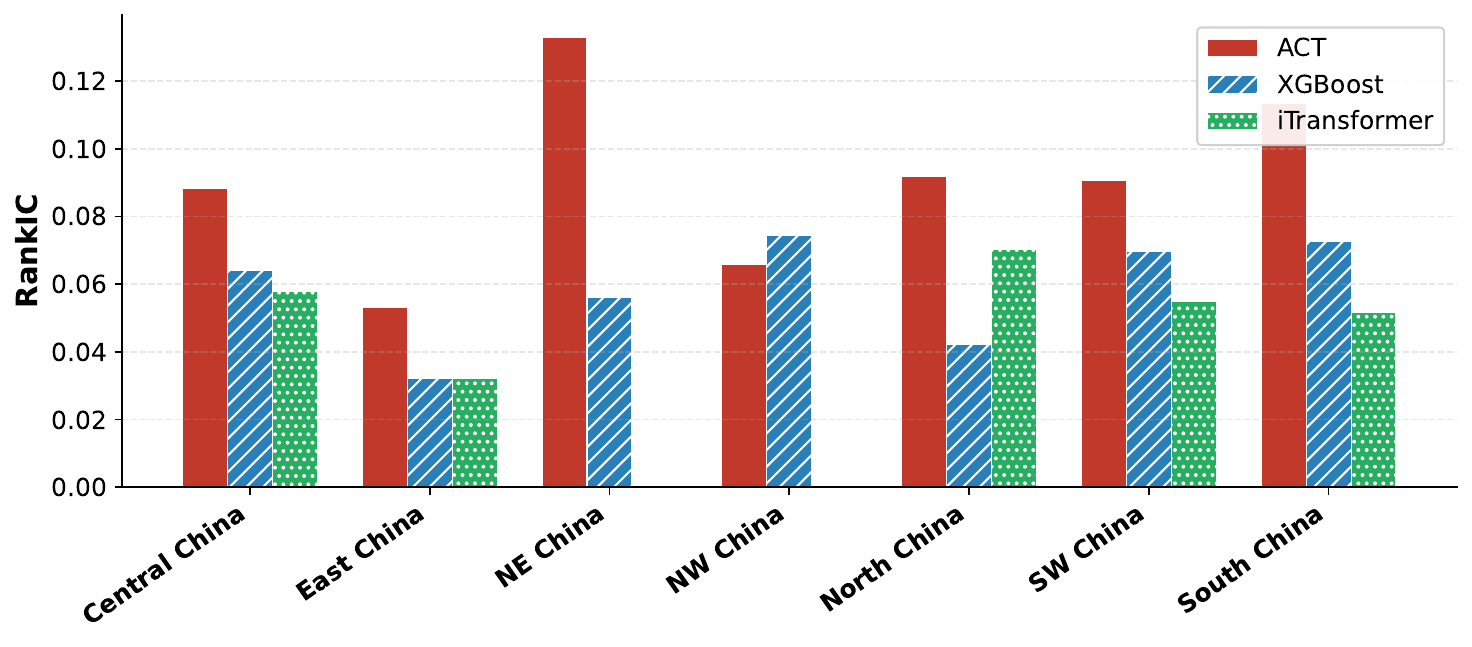}
        \caption{Rank IC by Region (CSI500)}
        \label{fig:appendix_bar_region_rankic500}
    \end{subfigure}%
    \hfill
    \begin{subfigure}[b]{0.48\textwidth}
        \centering
        \includegraphics[width=\textwidth]{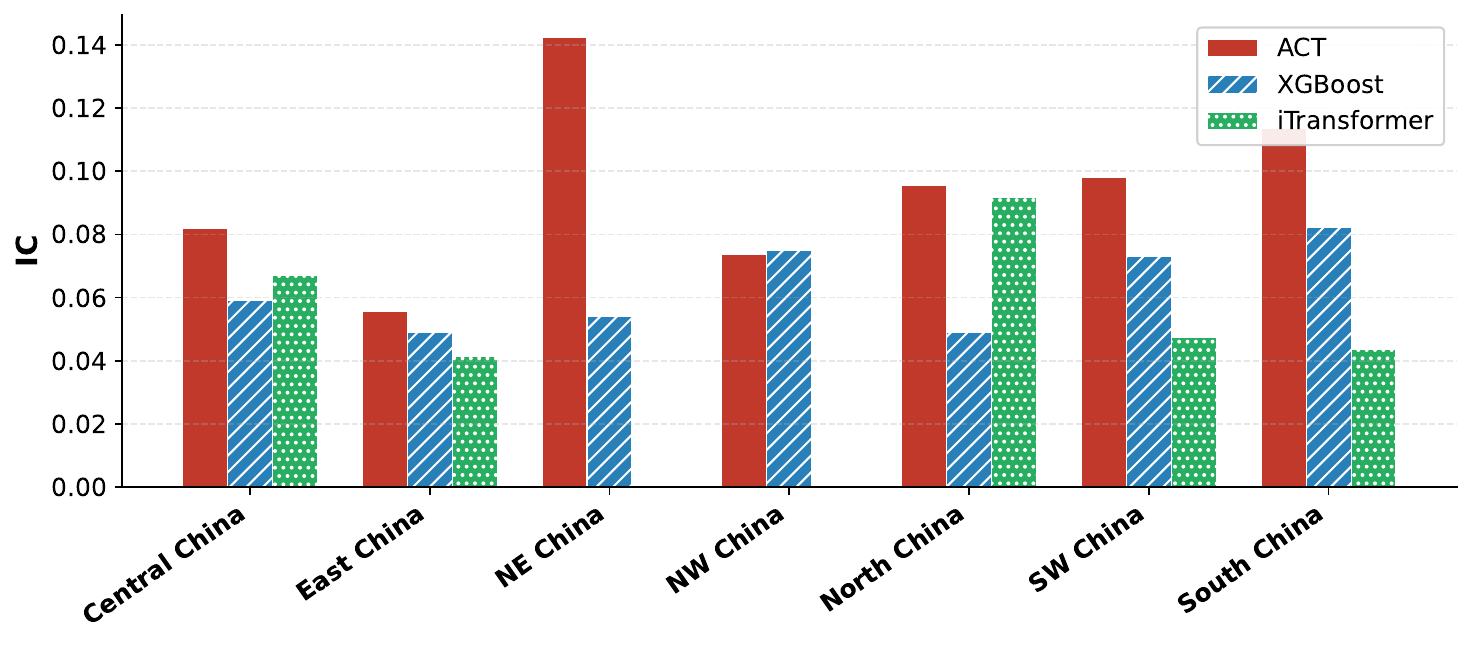}
        \caption{IC by Region (CSI500)}
        \label{fig:appendix_bar_region_ic500}
    \end{subfigure}
    \caption{Absolute IC and Rank IC comparison by geographic region on CSI500. Dense stock clusters (NE China, North China, South China) show clear ACT dominance; NW China is the sole exception. Companion to Figure~\ref{fig:region_heatmap}.}
    \label{fig:appendix_bar_region}
\end{figure*}

Figures~\ref{fig:subgroup_heatmaps}--\ref{fig:appendix_bar_region} and Tables~\ref{tab:subgroup_industry_csi500}--\ref{tab:subgroup_region_csi500} reveal a clear pattern that aligns with ACT's anti-crosstalk design. ACT achieves the highest RankIC in 11 of 15 sectors and the highest ICIR in 13 of 15 sectors. The largest gains appear in relationally dense sectors where inter-stock edges are plentiful and the PSPE module exerts its maximum effect: \emph{Pharma} (+78\% IC, +97\% ICIR)---573 listed companies with tight supply-chain and regulation linkages; \emph{Electronics} (+55\% IC, +87\% ICIR)---the largest sector with 1\,011 companies and highly interconnected upstream--downstream chains; and \emph{Retail} (+30\% IC, +56\% RankIC)---108 companies clustered in similar consumption channels.

Notably, even in sectors where iTransformer occasionally achieves a higher single-day IC than ACT (e.g.\ Construction $-28\%$ IC, Info.\ Tech.\ $-18\%$ IC, Materials $-16\%$ IC), ACT still leads substantially in ICIR and RankICIR (Info.\ Tech.\ +65\% RankICIR, Materials +50\% RankICIR), indicating that ACT's advantage lies in \emph{stable, day-to-day ranking quality} rather than sporadic peak correlation. Conversely, ACT trails across all four metrics only in \emph{Mining}---a sector with merely 52 companies and sparse relational structure---confirming that the anti-crosstalk mechanism requires sufficient relational density to be effective.

From the perspective of region relations, ACT leads all four metrics in 5 of 7 regions. The most striking case is \emph{NE China} (+164\% IC, +175\% ICIR, +137\% RankIC), where only 94 companies are listed but they concentrate heavily in a few industries (steel, heavy machinery), creating an extremely dense local graph.

Other dense clusters---\emph{South China} (767 companies in the Pearl River Delta manufacturing belt, +38\% IC, +56\% RankIC) and \emph{North China} (557 companies, +51\% ICIR, +86\% RankICIR)---also show consistent gains. The sole exception is \emph{NW China}, the sparsest region with only 114 companies spread across diverse industries; here ACT's RankIC trails the best baseline by $-12\%$, serving as a natural negative-control that reinforces the causal link between relational density and ACT's advantage.

\begin{table*}[h]
\centering
\caption{Subgroup analysis by industry sector on CSI500. \textcolor{red}{Red} values indicate the best model per sector. (Companion to Figures~\ref{fig:subgroup_heatmaps} and~\ref{fig:appendix_bar_industry}.)}
\label{tab:subgroup_industry_csi500}
\renewcommand{\arraystretch}{1.05}
\setlength{\tabcolsep}{3.5pt}
\footnotesize
\resizebox{\textwidth}{!}{
\begin{tabular}{lrrrrrrrrrrrr}
\toprule
\multirow{2}{*}{\textbf{Sector}} & \multicolumn{4}{c}{\textbf{ACT}} & \multicolumn{4}{c}{\textbf{XGBoost}} & \multicolumn{4}{c}{\textbf{iTransformer}} \\
\cmidrule(lr){2-5} \cmidrule(lr){6-9} \cmidrule(lr){10-13} 
 & IC & ICIR & RankIC & RankICIR & IC & ICIR & RankIC & RankICIR & IC & ICIR & RankIC & RankICIR \\
\midrule
Construction & 0.0740 & 0.1476 & \textcolor{red}{0.0709} & \textcolor{red}{0.1467} & -0.0120 & -0.0227 & -0.0098 & -0.0194 & \textcolor{red}{0.1022} & \textcolor{red}{0.1834} & 0.0587 & 0.1056 \\
Electronics & \textcolor{red}{0.1158} & \textcolor{red}{0.5638} & \textcolor{red}{0.1162} & \textcolor{red}{0.5762} & 0.0672 & 0.3013 & 0.0563 & 0.2707 & 0.0749 & 0.2361 & 0.0693 & 0.2314 \\
Equipment & \textcolor{red}{0.1018} & \textcolor{red}{0.4210} & \textcolor{red}{0.1011} & \textcolor{red}{0.4281} & 0.0719 & 0.2647 & 0.0602 & 0.2447 & 0.1011 & 0.2801 & 0.0993 & 0.2858 \\
Finance & 0.2073 & \textcolor{red}{0.6227} & \textcolor{red}{0.1985} & \textcolor{red}{0.6018} & 0.0890 & 0.2315 & 0.0663 & 0.1890 & \textcolor{red}{0.2097} & 0.4430 & 0.1822 & 0.3956 \\
Food \& Bev. & \textcolor{red}{0.1112} & \textcolor{red}{0.2923} & \textcolor{red}{0.0979} & \textcolor{red}{0.2609} & 0.0522 & 0.1349 & 0.0383 & 0.1053 & 0.0815 & 0.1704 & 0.0875 & 0.1885 \\
Info. Tech. & 0.1086 & \textcolor{red}{0.3033} & \textcolor{red}{0.1118} & \textcolor{red}{0.3211} & 0.0404 & 0.1074 & 0.0327 & 0.0931 & \textcolor{red}{0.1317} & 0.2292 & 0.1114 & 0.1941 \\
Materials & 0.0638 & \textcolor{red}{0.2612} & 0.0684 & \textcolor{red}{0.2855} & 0.0549 & 0.2031 & 0.0463 & 0.1903 & \textcolor{red}{0.0760} & 0.1938 & \textcolor{red}{0.0710} & 0.1833 \\
Mining & 0.0609 & 0.1505 & 0.0571 & 0.1406 & 0.0584 & 0.1385 & 0.0551 & 0.1370 & \textcolor{red}{0.0786} & \textcolor{red}{0.1853} & \textcolor{red}{0.0722} & \textcolor{red}{0.1683} \\
Other & \textcolor{red}{0.0957} & \textcolor{red}{0.3045} & \textcolor{red}{0.0936} & \textcolor{red}{0.2996} & 0.0531 & 0.1555 & 0.0485 & 0.1593 & 0.0888 & 0.2090 & 0.0616 & 0.1468 \\
Pharma & \textcolor{red}{0.0931} & \textcolor{red}{0.3993} & \textcolor{red}{0.0912} & \textcolor{red}{0.3895} & 0.0499 & 0.2029 & 0.0468 & 0.2090 & 0.0523 & 0.1452 & 0.0533 & 0.1546 \\
Real Estate & \textcolor{red}{0.0594} & \textcolor{red}{0.1288} & \textcolor{red}{0.0623} & \textcolor{red}{0.1335} & 0.0518 & 0.1021 & 0.0445 & 0.0930 & 0.0474 & 0.0899 & 0.0087 & 0.0166 \\
Retail & \textcolor{red}{0.1016} & \textcolor{red}{0.2462} & \textcolor{red}{0.0955} & \textcolor{red}{0.2354} & 0.0474 & 0.1048 & 0.0500 & 0.1220 & 0.0784 & 0.1910 & 0.0611 & 0.1463 \\
Textile \& Chem. & 0.0329 & 0.0654 & 0.0370 & 0.0716 & \textcolor{red}{0.0392} & \textcolor{red}{0.0728} & \textcolor{red}{0.0374} & \textcolor{red}{0.0724} & -- & -- & -- & -- \\
Transport & 0.0465 & 0.1041 & \textcolor{red}{0.0420} & \textcolor{red}{0.0941} & \textcolor{red}{0.0507} & \textcolor{red}{0.1107} & 0.0272 & 0.0641 & 0.0151 & 0.0280 & 0.0143 & 0.0263 \\
Utilities & 0.0459 & \textcolor{red}{0.1562} & 0.0485 & \textcolor{red}{0.1731} & 0.0481 & 0.1539 & 0.0471 & 0.1655 & \textcolor{red}{0.0578} & 0.1453 & \textcolor{red}{0.0570} & 0.1471 \\
\bottomrule
\end{tabular}}
\end{table*}

\begin{table*}[h]
\centering
\caption{Subgroup analysis by geographic region on CSI500. \textcolor{red}{Red} values indicate the best model per region.(Companion to Figure~\ref{fig:subgroup_heatmaps} and~\ref{fig:appendix_bar_region}.)}
\label{tab:subgroup_region_csi500}
\renewcommand{\arraystretch}{1.05}
\setlength{\tabcolsep}{3.5pt}
\footnotesize
\resizebox{\textwidth}{!}{
\begin{tabular}{lrrrrrrrrrrrr}
\toprule
\multirow{2}{*}{\textbf{Region}} & \multicolumn{4}{c}{\textbf{ACT}} & \multicolumn{4}{c}{\textbf{XGBoost}} & \multicolumn{4}{c}{\textbf{iTransformer}} \\
\cmidrule(lr){2-5} \cmidrule(lr){6-9} \cmidrule(lr){10-13} 
 & IC & ICIR & RankIC & RankICIR & IC & ICIR & RankIC & RankICIR & IC & ICIR & RankIC & RankICIR \\
\midrule
Central China & \textcolor{red}{0.0818} & \textcolor{red}{0.2344} & \textcolor{red}{0.0884} & \textcolor{red}{0.2563} & 0.0591 & 0.1602 & 0.0640 & 0.1879 & 0.0669 & 0.1423 & 0.0578 & 0.1269 \\
East China & \textcolor{red}{0.0557} & \textcolor{red}{0.2925} & \textcolor{red}{0.0531} & \textcolor{red}{0.2898} & 0.0491 & 0.2344 & 0.0320 & 0.1843 & 0.0415 & 0.1633 & 0.0322 & 0.1344 \\
NE China & \textcolor{red}{0.1423} & \textcolor{red}{0.2951} & \textcolor{red}{0.1329} & \textcolor{red}{0.2761} & 0.0539 & 0.1073 & 0.0560 & 0.1147 & -- & -- & -- & -- \\
NW China & 0.0736 & \textcolor{red}{0.1373} & 0.0658 & 0.1201 & \textcolor{red}{0.0750} & 0.1288 & \textcolor{red}{0.0744} & \textcolor{red}{0.1273} & -- & -- & -- & -- \\
North China & \textcolor{red}{0.0955} & \textcolor{red}{0.4173} & \textcolor{red}{0.0919} & \textcolor{red}{0.4086} & 0.0490 & 0.1911 & 0.0421 & 0.1814 & 0.0918 & 0.2766 & 0.0702 & 0.2200 \\
SW China & \textcolor{red}{0.0981} & \textcolor{red}{0.2602} & \textcolor{red}{0.0906} & \textcolor{red}{0.2533} & 0.0731 & 0.1784 & 0.0696 & 0.1928 & 0.0472 & 0.0915 & 0.0549 & 0.1110 \\
South China & \textcolor{red}{0.1134} & \textcolor{red}{0.4593} & \textcolor{red}{0.1134} & \textcolor{red}{0.4656} & 0.0823 & 0.2902 & 0.0727 & 0.2936 & 0.0436 & 0.0825 & 0.0516 & 0.0979 \\
\bottomrule
\end{tabular}}
\end{table*}

\end{document}